# Solution to Advanced Manufacturing Process Problems using Cohort Intelligence Algorithm with Improved Constraint Handling Approaches

Aniket Nargundkar[1*], Madhav Rawal[1], Aryaman Patel[1], Anand J Kulkarni[2], Apoorva S Shastri[2]

[1]Symbiosis Institute of Technology, Symbiosis International (Deemed University), Pune, India
[2]Institute of Artificial Intelligence, Dr Vishwanath Karad MIT World Peace University, Pune, India
Email: aniket.nargundkar@sitpune.edu.in, madhav.rawal.sitpune@gmail.com, aryaman.patel.sit@gmail.com, anand.j.kulkarni@mitwpu.edu.in, apoorva.shastri@mitwpu.edu.in

**Abstract**
Recently, various Artificial Intelligence (AI) based optimization metaheuristics are proposed and applied for a variety of problems. Cohort Intelligence (CI) algorithm is a socio inspired optimization technique which is successfully applied for solving several unconstrained & constrained real-world problems from the domains such as design, manufacturing, supply chain, healthcare, etc. Generally, real-world problems are constrained in nature. Even though most of the Evolutionary Algorithms (EAs) can efficiently solve unconstrained problems, their performance degenerates when the constraints are involved. In this paper, two novel constraint handling approaches based on modulus and hyperbolic tangent probability distributions are proposed. Constrained CI algorithm with constraint handling approaches based on triangular, modulus and hyperbolic tangent is presented and applied for optimizing advanced manufacturing processes such as Water Jet Machining (WJM), Abrasive Jet Machining (AJM), Ultrasonic Machining (USM) and Grinding process. The solutions obtained using proposed CI algorithm are compared with contemporary algorithms such as Genetic Algorithm, Simulated Annealing, Teaching Learning Based Optimization, etc. The proposed approaches achieved 2%-127% maximization of material removal rate satisfying hard constraints. As compared to the GA, CI with Hyperbolic tangent probability distribution achieved 15%, 2%, 2%, 127%, and 4% improvement in MRR for AJMB, AJMD, WJM, USM, and Grinding processes, respectively contributing to the productivity improvement. The contributions in this paper have opened several avenues for further applicability of the proposed constraint handling approaches for solving complex constrained problems.

**Article Highlights:**
1. Constraint handling approaches based on Modulus and Hyperbolic Tangent functions are developed.
2. Constrained version of Cohort Intelligence (CI) algorithm is presented with proposed constrained handling approaches.
3. Benchmark & Real-World Advanced Manufacturing Process problems are solved.
4. The proposed approaches achieved 1.6%-127% maximization of material removal rate satisfying hard constraints.

**Keywords:** Cohort Intelligence; Abrasive Jet Machining; Water Jet Machining; Ultrasonic Machining, Grinding Process, Constraint Handling Methods

## 1. Introduction & Motivation

Mathematical optimization or mathematical programming is the selection of the best element with respect to some criterion from a set of available alternatives (Du et al., 2008). It is the process of finding minimum and/or maximum states of the objective(s) from within many or possibly infinite solutions. For almost all the human activities there is a desire to deliver the most from the least. For example, in the business view point, maximum profit is desired with the least investment; maximizing the strength, durability, efficiency with minimum initial investment and operational cost of various industrial equipment and machineries. Thus, the concept of optimization carries great significance in both human affairs and the laws of nature, which is the inherent characteristic to achieve the best or most favourable (minimum or maximum) from a given situation (Kulkarni et al., 2015). While solving the real-world problems, there are certain practical restrictions referred to as constraints. They could be associated with certain resources such as cost, time and space limitation, design limitations, application specific requirements, etc. Designing optimal solution under all the diverse requirements/constraints is a challenging task (Arora, 2004). Traditionally, Exact methods have been applied to solve manufacturing problems. However, for complex NP hard problems, the exact methods could not find the optimal solution in the realistic time. This has led to the development of heuristic & metaheuristic algorithms. In the literature, it is suggested that the metaheuristics can efficiently solve large-scale non-linear models, models with large number of integer variables, models with not appropriately-defined objectives and constraints, and models with stochastic or dynamic elements (Jones & Tamiz, 2010). In the past, several nature-inspired/Artificial Intelligence (AI) based metaheuristics have been proposed which are broadly classified into bio-inspired, physics based and socio inspired methods. Some of the well-known examples of such nature-inspired metaheuristics are Genetic Algorithm (GA) (Goldberg, 1989), Particle Swarm Optimization (PSO) (Kennedy & Eberhart, 1995), Simulated Annealing (SA) (Van & Aarts, 1987), Black Hole Search algorithm (Hatamlou, 2013), Cuckoo Search algorithm (Gandomi et al., 2013) and Bio-geography based optimization (Simon, 2008). The socio-inspired optimization methods are the type of the AI based metaheuristics which are based on the competition and interaction of the societal individuals. The notable socio-inspired methods are Ideology Algorithm (Huan et al., 2017), Election Algorithm (Emami & Derakhshan 2015), The League Championship Algorithm (Kashan, 2009), Soccer League Competition Algorithm (Moosavian, 2015), Teaching Learning Based Optimization (Rao, 2016), Cultural Evolution Algorithm (Kuo and Lin, 2013), Social Learning Optimization (Liu et al., 2016), Anarchic Society Optimization (Ahmadi-Javid & Hooshangi-Tabrizi, 2017). An optimization approach Cohort intelligence (CI) based on societal artificial intelligence is developed by Kulkarni et al., 2013. Further, the variations of CI have also been proposed by Patankar and Kulkarni, 2018. In addition, Multi Cohort Intelligence (Multi-CI) algorithm developed by Shastri & Kulkarni, 2018. In addition, variations of CI and Multi-CI algorithms are applied for machining of Titanium alloy under Minimum Quantity Lubrication system (Shastri et al., 2019) and also for optimizing the Abrasive Water Jet Machining (AWJM) process (Gulia & Nargundkar, 2019).

Most of the real-world problems are constrained in nature viz. design constraints of a mechanical machine element, constraints for a manufacturing process, production constraints for a plant, time constraints for service, etc. Braune et al., 2004 used Genetic Algorithms (GA) to solve production planning problems of minimizing the mean deviance from the customer orders' predetermined delivery dates, mean earliness & tardiness and maximizing the workplace utilization. Ponsich et al., 2008 used GA for optimal batch plant design involving time constraint. Askarzadeh et al., 2016 proposed Crow Search Algorithm to solve constrained engineering optimization problems viz. three bar truss design, pressure vessel problem, welded beam design. Rao et. al., 2011 applied Teaching Learning Based Optimization Algorithm (TLBO) to optimize the parameters of advanced machining processes such as Electrochemical Machining (ECM) & Electrochemical Discharge Machining (EDM). Mallipeddi & Suganthan, 2010 hybridized four constraint handling techniques with Evolutionary Algorithms. The results of the research showed that combining different constraint handling techniques when used with evolutionary programming can outperform single constraint handling methods. Biswas et. al., 2018 solved non-linear Power Flow optimization problem using Differential Evolution (DE) algorithm. Zhao et al., 2022 proposed and applied a novel convolutional deep belief network with Gaussian distribution for bearing fault diagnosis and classifications. Song et al., 2023 proposed a co-evolutionary multi-swarm adaptive differential evolution algorithm to solve the premature convergence and search stagnation challenges of a classical DE algorithm and applied it for portfolio optimization problem. Chen et al., 2023 presented a novel HSI classification network called MS-RPNet, viz., multiscale superpixelwise RPNet and applied for image processing problems. Li et al., 2023 designed a blockchain-based flight operation data sharing scheme, named BFOD to achieve the privacy protection and secure sharing of flight operation data.

Any metaheuristics is initially developed for solving unconstrained problem and further modified with constraint handling technique for solving constrained optimization problems. It is observed that, the performance of the algorithm is affected after incorporating the constraint handling approaches. So far, various approaches have been developed and applied with EAs. Constraint handling techniques are classified by Lagaros et al., 2023 in categories viz. penalty-based techniques, repair algorithm-based techniques, boundary-based techniques, etc. Four of the most commonly employed constraint handling technique formulations are the penalty methods, feasibility rules, $\varepsilon$-constrained method, and stochastic ranking. The penalty based approaches are widely applied in the literature owing to the ability of transforming constrained problem into unconstrained one without adding much complexity (He et al., 2019, Deb, 2012). Several penalty-based approaches are developed so far such as static penalty approach (Homaifar et al., 1994), dynamic penalty approach (Joines & Houck, 1994), annealing penalty approach (Michalewicz et al.,1994), exact penalty function method (Huyer et al., 2003), barrier function methods (Arora, 2004), etc. However, literature reports that the Penalty function constraint handling methods susceptible to penalty parameters and often are problem specific. For the feasibility based constraint handling approaches, a mechanism needs to be incorporated to push the solution towards feasible region increasing the computational cost. Kulkarni & Shabir, 2016 proposed a triangular probability based constrained handling approach by exploiting this inherent capability of CI. CI algorithm using static and dynamic penalty function based approaches was proposed by Kulkarni et al., 2018. The constraint handling ability of CI was investigated with static penalty and dynamic penalty approach. In static penalty approach, infeasible solutions are penalized with a constant penalty. The pseudo-objective function is formulated considering penalty parameter for the violation of the equality and inequality constraints. For the dynamic penalty approach, a minimal penalty is imposed initially for infeasible solutions and as the algorithm progresses, penalty value is increased for every learning attempt. This constrained CI algorithm was validated by solving benchmark problems and further applied on real world constrained mechanical engineering design problems. Furthermore, Kale & Kulkarni, 2021 proposed a self-adaptive penalty function (SAPF) approach referred to as CI–SAPF. Also, the approach is Hybridized with Colliding Bodies Optimization (CBO) algorithm to develop CI-SAPF-CBO algorithm. The CI algorithm incorporating a probability based constraint handling approach was proposed by Shastri et al, 2016. In this approach a probability distribution was devised for every constraint. The lower and upper bounds of the distribution are chosen by finding the minimum and maximum values among all the constraints. Based on the range in which the value of each of the constraint lies, the probability and the probability scores were calculated. Roulette wheel approach based on the probability score is used to select the behavior to be followed, and range reduction approach is used to arrive at the best solution. The mechanical engineering design optimization problems were solved using CI algorithm with the probability based constraint handling approach by Shastri et al., 2019. In the current work, novel constraint handling approaches are proposed and constrained CI algorithm is presented.

The 20th century has seen the introduction of new generation of materials and alloys comprising of special characteristics such as high strength-to-weight ratio, high stiffness and toughness, high heat capacity and thermal conductivity, etc. due to the demand from various applications. The machining of advanced materials has brought new challenges such as rapid deterioration of the cutting tools, inferior quality of machined parts, etc. The major reasons behind these problems are generation of high temperatures and stresses during machining of these advanced materials (Shastri et al., 2020, Shastri et al., 2021). This has led to the development and establishment of the Non Traditional Machining (NTM) or Advanced Manufacturing Processes (AMP) as efficient and economic alternatives to the conventional ones. Today, NTMs/ AMPs with varied machining capabilities and specifications are available for a wide range of applications. The AJM, WJM, USM, EDM, Laser Beam Machining (LBM) and Grinding are some of the examples of the AMPs. Today's manufacturing industry is trying to address the challenges such as growing needs for safety, reduced time-to-market that implies short manufacturing time, minimal manufacturing costs through the efficient use of the resources and, expected quality of highly customized products. Determining optimal process parameter settings critically influences productivity, quality, and cost of production. Therefore, optimal manufacturing process parameter setting is recognized as one of the most important activity (Orio et al., 2013, Pansari et al., 2019).

Several studies have been performed to optimize the process parameters for AMPs. In recent times, owing to the increase in the kinds of hybrid materials and polymers used in the aerospace industry, AJM process is becoming widely used to machine materials like particle reinforced aluminium alloy, CFRP, titanium, Inconel (Rao et al., 2014). Tomy et al., 2020 investigated the abrasive jet machining process for hybrid silica glass fibre reinforced composites. Prasad et al., 2018 performed parametric optimization using WASPAS and MOORA to optimize different interconnected responses during the AJM process. Kumar et al., 2020 reviewed the methods used to optimise the parameters of the AJM process. The optimization problem of the AJM considered in this paper, aims at maximizing the Material Removal Rate (MRR) while keeping a constraint of surface roughness as a measure of surface finish. The decision variables that affect the WJM process are the nozzle diameter, transverse rate, stand-off distance and the jet pressure on nozzle exit. These variables affect the objective of maximizing the MRR and minimizing the specific energy. Through experimentation the constraints on the system are considered to be the process parameters along with power consumption which is a function of the pressure at nozzle exit and the nozzle diameter (Jain et. al., 2007, Mellal & Williams 2016). The water pressure on nozzle exit directly affects the MRR (linearly dependent). The ability of the water to cut into different materials is dependent on the pressure thresholds of the said materials. The transverse feed rate is controlled between ranges to ensure increased MRR. It has a lower range (2-4m/s) for cutting metal however, with the further increase in feed rate (4-8m/s) can reduce the MRR (Mishra et. al., 2012). Biswas et al., 2019 studied the effect of USM parameters on several factors like MRR, taper angle and overcut during machining of zirconia composite. Das et al., 2013 investigated surface roughness and MRR of USM in machining of hexagonal holes in zirconia bio-ceramics. Goswami et al., 2015 used gravitational search algorithm and fireworks algorithm for parametric optimisation of USM and compared it with other popular population - based algorithms. Furthermore, modelling and optimization of systems are achieved with various contemporary algorithms. Adaptive Dynamic Programming (ADP) based control algorithm is applied for solving dynamic optimization of Hydraulic Servo Actuators (HSA) (Djordjevic et al., 2023). A self-triggered Model Predictive Control (MPC) approach is developed and applied for discrete-time semi-Markov jump linear systems to achieve a desired finite-time performance (He et al., 2022). A hidden Markov model is developed and applied for control of a wind turbine system (Cheng et al., 2022).

Authors have earlier applied recently developed socio-inspired optimization method referred to as multi-cohort intelligence (Multi-CI) for solving real-world AMP optimization problems. The problems considered were minimization of surface roughness for AWJM, EDM, micro-turning and micro-milling processes. Furthermore, the taper angle for the AWJM, relative electrode wear rate for EDM, burr height and burr thickness for micro-drilling, flank wear for micro-turning process, machining time for micro-milling processes were minimized (Shastri et al., 2020). However, all these were unconstrained problems. In the current work, proposed constrained CI algorithm is applied for real world constrained problems from the advanced manufacturing domain such as AJM, WJM, USM and Grinding process. The problems solved are maximizing Material Removal Rate ($MRR$) for USM and AJM with brittle & ductile materials considering the surface roughness $R_a$ constraint. Furthermore, maximizing ($MRR$) for WJM & Grinding with constraints such as power usage for WJM and $R_a$ & No of Flaws for Grinding processes. All the problems are referred to from Mellal & Williams, 2016.

As stated earlier, the CI algorithm with triangular probability distribution aims to maximize the probability of candidate and hence the constraints are forced to come close to zero and be satisfied. However, in the positive neighborhood of lower bound and the negative neighborhood of upper bound, the probability of the constraint values outside the bounds is more than that at the bounds. This makes the points outside the constraint bounds seem favorable which may cause the algorithm to go awry. Moreover, the points closer to the boundary but outside the triangle and the points far away from the triangle, both are assigned the same amount of penalty. This reduces the exploration capabilities of the algorithm as it forces the values in the triangle. The resulting values may satisfy constraints but could be local optima. Hence, there is a need of an alternative algorithm that tries to navigate through these problems and present better results with complete exploration and exploitation capabilities. In the current work, constrained CI algorithm with three constraint handling approaches based on triangular, modulus & hyperbolic tangent probability distributions is proposed. The proposed approaches are validated by solving constrained test problems G1, G4 and G6. These problems are selected because the global optimum function value for these problems is negative and hence these problems are best suited for investigating the exploration and exploitation capabilities of algorithm through positive and negative search space. Further, the applicability of proposed approaches is validated by solving real world problems from advanced manufacturing processes viz. Abrasive Jet Machining (AJM), Water Jet Machining (WJM), Ultrasonic Machining (USM) and Grinding process.

The novelty and contribution of the current work is as follows:
1. Constraint handling approaches based on Modulus and Hyperbolic Tangent functions are developed.
2. Constrained Cohort Intelligence (CI) algorithm with proposed constraint handling approaches is presented.
3. Benchmark & real-world advanced manufacturing process problems are solved using proposed CI algorithm.
4. The real-world constrained Water Jet Machining (WJM), Abrasive Jet Machining (AJM), Ultrasonic Machining (USM) and Grinding process problems are solved and as compared to the GA, CI with Hyperbolic tangent probability distribution achieved 15%, 2%, 2%, 127%, and 4% improvement in MRR for AJMB, AJMD, WJM, USM, and Grinding processes, respectively.
5. The robustness of the proposed CI algorithm has been established for solving the real-world problems.
6. Applicability of proposed CI algorithm for solving constrained advanced manufacturing problems is successfully demonstrated.

The remainder of the paper is organized as follows: Section 2 discusses the constrained CI algorithm with the proposed constrained handling methods in detail. Section 3 describes the AJM, WJM, USM and Grinding problems. It includes the importance of the problems, mathematical formulations. It is followed by the discussion on the results and comparison amongst several algorithms in Section 4. The conclusions, notable contributions and a comment on future directions are provided at the end in Section 5.

## 2. Proposed Approaches

This section describes the methodology adopted in the current work for optimizing AMP parameters. Section 2.1 presents the triangular probability distribution method, whereas Section 2.2 discusses the drawbacks of this approach. The two proposed probability distribution approaches referred to as modulus & hyperbolic tangent are illustrated in Sections 2.3 & 2.4, respectively.

As discussed in Section 1, CI algorithm models the ability of candidates in a cohort to self-supervise and improve their independent behavior. It is based on the natural tendency of an individual to evolve its behavior by observing the behavior of other candidates of the cohort and emulating it. Every candidate follows a certain behavior and the associated qualities which may improve its own behavior. In order to arrive at the best solution, every candidate shrinks the sampling interval associated with every variable by using a reduction factor which along with the number of candidates is determined based on preliminary trials. The cohort behavior is considered to be saturated/converged if further significant improvement in the behavior is not possible. The pseudo code for the CI algorithm is presented in the Figure 1. For details of the mathematical formulations and flowchart of CI refer to the appendix of Kulkarni et al., 2016.

```
Objective function Minimize f(x)
Select number of candidates in the cohort c
Set interval reduction factor r
Set convergence parameter ϵ
While (No of Learning attempts < Max Learning attempts)
    Generate the Randomize qualities of each candidate
    Evaluate objective function for every candidate.
    Evaluate probability associated for every candidate in the cohort
    Use following strategy (RW, FBest, FBetter, or Alienation) to select behaviour to follow by each candidate.
    Shrinkage of the interval for every candidate
    If (Convergence criterion met)
        Accept the current best candidate and its behaviour as a final solution
    Else
        Generate the Randomize qualities of each candidate as 2nd iteration
    End
End
```

**Figure 1: Pseudo Code of CI Algorithm**

**2.1 CI with a Triangular Probability Distribution Method**
In the probability distribution method, probability of the selection of the candidate is distributed between two fixed values of the constraints ($k_l$ & $k_u$). If the constraint values lie in between the specified range, probability is mapped with a linear function, otherwise it is given a negligible value, indicating that the candidate is outside the feasible region and shows lesser probability of survival. The Triangular Probability Distribution is presented in Figure 2.

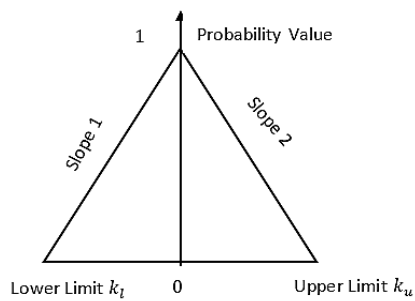

**Figure 2: Triangular Probability Distribution**

**2.2 Drawback of the Triangular Probability distribution function: The need of a new approach.**
In the existing approach, a graph plane with constraint values on the x axis and probability on the y axis is plotted as shown in Figure 3. A lower limit and an upper limit of constraints is selected and lines are drawn from this point to (0,1) as the maximum probability of any event is 1. The equations of these lines are calculated and the probability associated to constraint values within the bounds is calculated with the equation of the lines. If the constraint values lie outside the specified bounds, an arbitrary value of probability is assigned (viz. in the order of 1e-4). This is shown in the Figure 3.

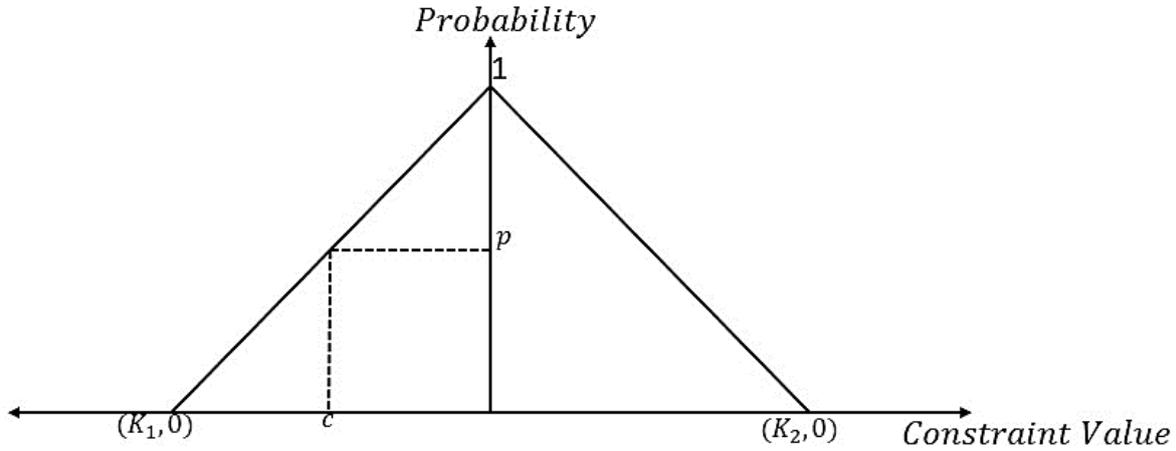

**Figure 3: Drawbacks Triangular Probability Distribution**

The algorithm aims to maximize the probability of survival of the $i^{th}$ candidate given by Eq. 1.

$$probability = \frac{p_i}{\sum_{j=1}^{n} p_j} \qquad (1)$$

and hence the constraints are forced to come close to 0 and be satisfied. However, in the positive neighborhood of lower bound and the negative neighborhood of upper bound, the value of probability tends to 0. This is because at boundary points, the line intersects the x axis and hence the y coordinate (probability) drops to 0. At these points, the probability of the constraint values outside the bounds is more than that at the bounds. This makes the points outside the constraint bounds seem favorable which may cause the algorithm to go awry. Moreover, the points closer to the boundary but outside the triangle and the points far away from the triangle, both are assigned the same amount of penalty. This reduces the exploration capabilities of the algorithm as it forces the values in the triangle. The resulting values may satisfy constraints but could be local optima. Therefore, there is a need of an alternative algorithm that tries to navigate through these problems and present better results with complete exploration and exploitation capabilities.

**2.3 Modulus Distribution Function**
The first approach to counter the situation mentioned in Section 2.2 is intuitive. The problem is at the intersection of the lines with the x axis hence, consider the mirror image of the graph: a modulus function. A modulus function of the form as shown in Eq. 2.

$$y = |ax| + \delta \qquad (2)$$

is considered where $a$ is an arbitrary constant and $\delta$ is an infinitesimal tolerance value. Instead of a probability approach, a penalty approach is considered where if the constraints are out of the bounds $[k_1, k_2]$, a penalty of the form of $y = \varphi|ax|$ is imposed where $\varphi$ is the static penalty.

If the value is inside the bounds, a mild penalty is calculated as shown in Figure 4. Unlike the Triangular distribution, here the goal of the algorithm is to minimize the penalty as the equation converges to 0. The probability of survival based on constraint for the $i^{th}$ candidate is calculated as shown in Eq. 3.

$$probability = \frac{\frac{1}{p_i}}{\sum_{j=1}^{n} \frac{1}{p_j}} \qquad (3)$$

However, this approach is not without drawbacks. Depending upon the optimization problem, if the values of constraints are very high, applying the penalty function may increase the computation cost. This causes the need of an improved function which circumnavigates the concept of penalty.

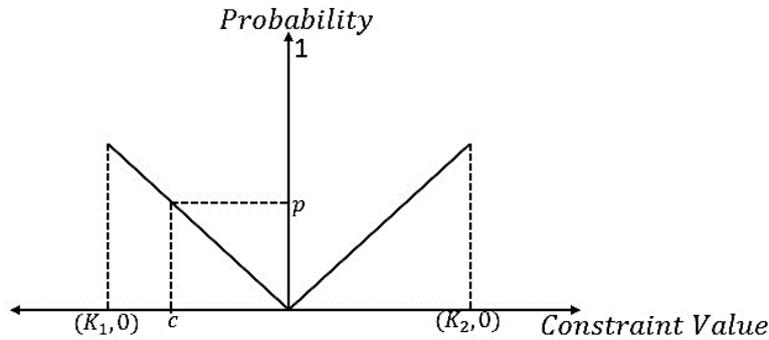

**Figure 4: Modulus Probability Distribution function**

**2.4 Hyperbolic Tangent Probability Distribution Function**

The mathematical function of hyperbolic tangent as shown in Figure 5 finds its application in deep learning algorithms as an alternative to the famous sigmoid function which is widely used in classification algorithms.

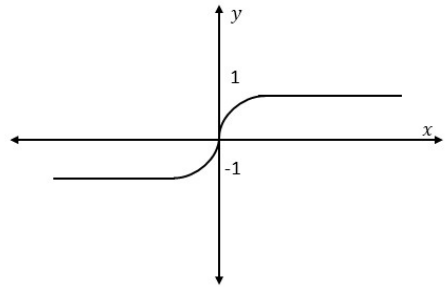

**Figure 5: Hyperbolic Tangent function**

The range of this function is [-1, 1]. The generalized form of this function is given in Eq. 4.

$$y = tan\,h(a|x|) + \delta \tag{4}$$

In this method $\delta$ is an arbitrary infinitesimal tolerance and the constant $a$ is calculated as per the required bounds $k$ as shown in Eq. 5.

$$a = \frac{tanh^{-1}(0.999)}{k} \tag{5}$$

The aim of this algorithm is to minimize probability so that the constraints are forced to zero as shown in Figure 6. The $tanh$ function will assign a high probability ($\sim$1) to constraints that are outside the specified bounds and therefore will have a lower probability for survival. The tolerance will prevent the discontinuity of the probability function near the origin as the probability for survival is calculated as presented in Eq. 6.

$$probability = \frac{\frac{1}{p_i}}{\sum_{j=1}^{n}\frac{1}{p_j}} \tag{6}$$

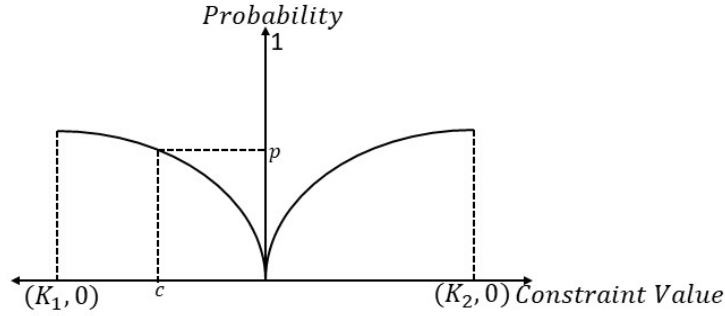

Figure 6: Hyperbolic Tangent Probability Distribution function

## 3. Problem Description
This section describes the problem formulations considered for the current work. Section 3.1 and 3.2 presents the AJM and WJM problems, respectively. The problems of USM and Grinding are discussed in Section 3.3 and 3.4, respectively.

### 3.1 Abrasive Jet Machining
In this machining process a jet of abrasive particles with carrier gas are released from a nozzle. Particles has very high velocities, impinge the target surface and remove the material by erosion. It is a highly efficient process for ductile and brittle metals and alloys, ceramics, and semiconductors (Jain et al., 2007). The model of AJM for brittle (Eq. 7&8) and ductile materials (Eq. 9&10) by Mellal & Williams,2016 for maximization of $MRR$ with surface roughness constraint along with three decision variables such as mass flow rate of abrasive particles $M_a$ (kg/s), mean radius of abrasive particles $r_m$ (mm), and velocity of abrasive particles $v_a$ (mm/s) is used. Table 1 and 2 provides the values of constants used in Eq. 8 & 10, respectively.

#### 3.1.1 Brittle Materials:

$$\text{Maximize Material Removal Rate} = 0.0035 \frac{\eta_a}{\sigma_{fw}^{0.75} \rho_a^{0.25}} M_a v_a^{1.5} \tag{7}$$

$$\text{Surface Roughness Constraint} = \frac{18.26}{(Ra)_{max}} \left[\frac{\rho_a}{\sigma_{fw}}\right]^{0.5} r_m v_a - 1 \leq 0 \tag{8}$$

With limits

$$0.0000167(kg/s) \leq M_a \leq 0.0005(kg/s)$$

$$0.005(mm) \leq r_m \leq 0.075(mm)$$

$$150{,}000(mm/s) \leq v_a \leq 400{,}000(mm/s)$$

Table 1: Constants table for Abrasive Water Jet Machining for Brittle Materials (AWJMB)

| Constants notation | Details | Units | Value |
|---|---|---|---|
| $\eta_a$ | Proportion of abrasive particles effectively participating in | NA | 0.7 |
| $\rho_a$ | Density of abrasive particles | Kg/mm³ | 3.85x 10⁻⁶ |
| $(Ra)_{max}$ | Allowable surface roughness value | μm | 0.8 |
| $\sigma_{fw}$ | Flow stress of work material | MPa | 5000 |

Note: NA- Not Applicable

#### 3.1.2 Ductile Materials

$$\text{Maximize Material Removal Rate} = 1.0436 \times 10^{-6} \zeta \left(\frac{\rho_w}{\delta_{cw}^2 H_{dw}^{1.5} \rho_a^{0.5}} M_a v_a^3\right) \tag{9}$$

**8**

$$\text{Surface Roughness Constraint} = \frac{25.82}{(R_a)_{max}} \left[\frac{\rho_a}{H_{dw}}\right]^{0.5} r_m v_a - 1 \leq 0 \quad (10)$$

With Limits =
$$0.0000167(kg/s) \leq M_a \leq 0.0005(kg/s)$$
$$0.005(mm) \leq r_m \leq 0.075(mm)$$
$$150{,}000(mm/s) \leq v_a \leq 400{,}000(mm/s)$$

**Table 2: Constants table for Abrasive Water Jet Machining for Ductile Materials (AWJMD)**

| Constants notation | Details | Units | Value |
|---|---|---|---|
| $\rho_w$ | Density of work material | Kg/mm³ | 2.7x 10⁻⁶ |
| $H_{dw}$ | Dynamic hardness of work material | MPa | 1.15 |
| $(R_a)_{max}$ | Allowable surface roughness value | μm | 2 |
| $\rho_a$ | Density of abrasive particles | Kg/mm³ | 2.48x 10⁻⁶ |
| $\delta_{cw}$ | Critical plastic strain or erosion ductility of work material | NA | 1.5 |
| $Z$ | Amount of indentation Volume plastically-deformed | NA | 1.6 |

Note: NA- Not Applicable

### 3.2 Water Jet Machining:

Water Jet machining is a non-traditional advance machining process that utilizes a hydraulically pressurized water jet to bore into or make precision cutting on workpieces. We have used the model of WJM by Mellal & Williams,2016 for maximization of $MRR$ with power usage constraint along with four decision variables such as the nozzle diameter, transverse rate, stand-off distance and the jet pressure on nozzle exit. The objective function is described in eq. 11 whereas Eq. 12- 14 presents the calculation of intermediate terms in Eq.11. The power usage constraint is shown in Eq.15. Table 3 provides the values of constants used in Eq. 11 & 15.

$$\text{Maximize Material Removal Rate} = \frac{0.297}{C_{fw}} d_{wn}^{1.5} f_n X^{0.5} \psi^{\left(\frac{2}{3}\right)} \left[1 - \frac{\sigma_{yw}}{2P_w\phi}\right] \left[1 - e^{\left(-2256.76\frac{C_{fw}P_w\phi}{\eta_w f_n}\right)}\right] \quad (11)$$

Where:

$$\phi = \frac{2}{K}[0.5 - 0.57\psi + 0.2\psi^2] \quad (12)$$

$$\psi = \left(1 - \sqrt{\frac{1}{P_w}\frac{\sigma_{pw}K}{2}}\right) \quad (13)$$

$$K = \frac{X_i}{X} \quad (14)$$

Subjected to Power Usage limit =

$$\frac{0.777 \times 10^{-1.5} d_{wn}^2 P_w^{1.5}}{P_{max}} - 1 \leq 0 \quad (15)$$

With limits:
$$1(MPa) \leq P_w \leq 400(MPa)$$
$$0.05(mm) \leq d_{wn} \leq 0.5(mm)$$
$$1(mm/s) \leq f_n \leq 300(mm/s)$$
$$2.5(mm) \leq X \leq 50(mm)$$

**Table 3: Constants table for Water Jet Machining (WJM)**

| Constants notation | Details | Units | Value |
|---|---|---|---|
| $P_{max}$ | Allowable power consumption | kW | 50 |
| $\sigma_{pw}$ | compressive yield strength of work material | MPa | 26.2 |
| $C_{fw}$ | Drag friction coefficient for work material | NA | 0.005 |
| $\sigma_{yw}$ | tensile yield strength of the work material | MPa | 3.9 |

| $\eta_a$ | Damping coefficient of the work material | Kg mm⁻²s⁻¹ | 2357.3 |
| $X_i$ | Length of initial region of water jet | mm | 20 |

Note: NA- Not Applicable

## 3.3 Ultrasonic Machining

Ultrasonic machining is a non-conventional machining process that is used to machine materials with high hardness and low ductility. We have used the model of WJM by Mellal & Williams, 2016 for maximization of MRR with surface roughness constraint along with five decision variables such as frequency of vibration $f_v$ (Hz), amplitude of vibration $A_v$ (mm), mean diameter of abrasive grain $d_m$ (mm), static feed force. $F_s$ (N), and Volumetric concentration of abrasive particles in slurry $C_{av}$. The objective function is described in eq. 16. The surface roughness constraint is shown in Eq.17. Table 4 provides the values of constants used in Eq. 16 & 17.

$$\text{Maximize Material Removal Rate} = \frac{4.963 A_t^{0.25} K_{usm}^{0.75}}{[\sigma_{fw}(1+\lambda)]^{0.75}} C_{av}^{0.25} F_s^{0.75} A_v^{0.75} d_m f_m \quad (16)$$

$$\text{Surface Roughness Constraint} = \frac{1154.7}{[A_t \sigma_{fw}(1+\lambda)]^{0.5} (R_a)_{max}} \left[\frac{F_s A_v d_m}{C_{av}}\right]^{0.5} - 1 \leq 0 \quad (17)$$

With Limits

$$0.005 (mm) \leq A_v \leq 0.1 (mm)$$
$$10{,}000 (Hz) \leq f_v \leq 40{,}000 (Hz)$$
$$0.007 (mm) \leq d_m \leq 0.15 (mm)$$
$$0.05 \leq C_{av} \leq 0.5$$
$$4.5 (N) \leq F_s \leq 45 (N)$$

**Table 4: Constants table for Ultrasonic machining**

| Constants notation | Details | Units | Value |
|---|---|---|---|
| $A_t$ | Cross-sectional area of cutting tool | mm² | 20 |
| $(R_a)_{max}$ | Allowable surface roughness value | μm | 0.8 |
| $\sigma_{fw}$ | Flow stress of work particle | MPa | 6900 |
| $K_{usm}$ | A constant of proportionality relating mean diameter of abrasive | mm⁻¹ | 0.1 |
| $\sigma_{ft}$ | Flow stress of abrasive particle | MPa | 28000 |

## 3.4 Grinding Process:

Grinding Process is a finishing operation that utilizes an abrasive wheel to chip off material from workpiece. This machining process is becoming of rapid importance in automobile and aerospace industries (Jackson et. al. 2011). The model of WJM by Mellal & Williams, 2016 for maximization of MRR with surface roughness & no. of flaws constraints along with three decision variables such as feed rate, depth of cut and grit size is used. The objective function is described in eq. 18. The surface roughness constraint is shown in Eq.19 whereas the constraint regarding no. of flaws is presented in Eq. 20.

$$\text{Maximize Material Removal Rate} = (f_r) * (d_c) \quad (18)$$

$$\text{Surface Roughness constraint } SR_{max} (\mu m) = 0.145(d_c)^{0.1939}(f_r)^{0.7071}(M)^{-0.2343} \leq 0.3 \quad (19)$$

$$\text{Number of Flaws } ND_{max} = 29.67(d_c)^{0.4167}(f_r)^{0.8333} \leq 7 \quad (20)$$

With limits:

$$0.86 (m/min) \leq f_r \leq 13.4 (m/min)$$
$$5 (\mu m) \leq d_c \leq 30 (\mu m)$$

$$120 \leq M \leq 500$$

## 4. Results & Discussions

In this section, validation of the proposed approaches by solving constrained test problems is presented. Furthermore, a real world application of the proposed approaches is demonstrated by applying it to five AMP parameters optimization problems. As mentioned earlier, all the problems are referred from Mellal & Williams, 2016. The results are compared with various contemporary algorithms like TLBO, GA, SA, Hoopoe heuristic and cuckoo optimisation. Every problem is solved for 30 times and the Standard Deviation (SD) is also presented for every problem. A vectorized coding approach has been implemented in the algorithm as opposed to the conventional loops which give a huge boost to decrease the computation time. The vectorized CI algorithm along with the 3 constraint handling techniques are coded in MATLAB R2020a on Windows Platform with an Intel Core i5 processor and 8 GB RAM. Table 5 presents the control parameters set for solving all the problems considered in the current work. Initially, the default setting of parameters are used. Different values of control parameters are later explored, and no significant improvement in results have been observed. Hence, these parameters are selected for the experimentation in the current work.

**Table 5: Parameters used in the problems solved**

| Problem | No. of Candidates | Parameter | | Reduction Factor (r) | Convergence |
|---|---|---|---|---|---|
| | | k1 | k2 | | |
| Abrasive Jet (Brittle) | 5 | -10 | 1 | 0.99 | Range 1e-15 |
| Abrasive Jet (Ductile) | 5 | -1 | 1.5 | 0.99 | Range 1e-15 |
| Ultrasonic Machining | 5 | -10 | 1 | 0.99 | Range 1e-15 |
| Grinding Process | 5 | -100 | 5 | 0.99 | Range 1e-15 |
| Water Jet Machining | 5 | -1 | 1 | 0.99 | Range 1e-15 |

### 4.1 Validation of the proposed approaches

In this section, the proposed hyperbolic tangent probability distribution and Modulus Distribution with CI approaches are validated by solving Global Standard Test suit problems. Tables 6 and 7 presents the results obtained for Global Constrained Test Problems (G1, G4, G6) with hyperbolic tangent probability distribution and Modulus Distribution, respectively.

**Table 6: Results for CI with hyperbolic tangent Probability Distribution**

| Problem | Global | Function Value | Mean | SD | No. of | Time |
|---|---|---|---|---|---|---|
| G1 | -15.0000 | -15.0300 | -15.0290 | 3.65E-4 | 895 | 0.1663 |
| G4 | -30665.5390 | -30665.4900 | -0664.9220 | 2.84E-5 | 954 | 0.1445 |
| G6 | -6961.8130 | -6961.8290 | -6961.8190 | 1.43E-4 | 912 | 0.0884 |

**Table 7: Results for CI with Modulus Probability Distribution**

| Problem | Global | Function Value | Mean | SD | No. of | Time |
|---|---|---|---|---|---|---|
| G1 | -15.0000 | -15.0200 | -15.0280 | 1.85E-5 | 981 | 0.0845 |
| G4 | -30665.5390 | -30664.8740 | -0663.1320 | 6.44E-3 | 1002 | 0.6642 |
| G6 | -6961.8130 | -6962.7410 | -6961.6820 | 4.13E-5 | 872 | 0.0412 |

### 4.2 Solutions to Real-World Problems

This section describes five real world Problems which are solved using the Triangular, Hyperbolic tangent and Modulus Probability Distributions with CI. The results obtained by the CI algorithm are also compared with the results described in Mellal & Williams, 2016 which includes solutions from various popular algorithms like TLBO, GA, SA, Hoopoe heuristic and cuckoo optimisation. Tables 8, 9 and 10 presents results for all the problems in the current work such as AJM for brittle & ductile materials, WJM, USM and Grinding, solved using Triangular, Modulus and Hyperbolic tangent Probability Distributions, respectively.

**11**

**Table 8: Results for all problems solved using Triangular Probability Distribution**

| Problem | Function Value | Constraint Value | Mean | Best | SD | No. of iterations | Time in Seconds |
|---|---|---|---|---|---|---|---|
| AJMB | 8.2528 | -4.44E-16 | 8.2456 | 8.2528 | 0.0285 | 1342 | 0.1408 |
| AJMD | 0.6056 | 1.11E-16 | 108.1131 | 0.6056 | 0.0081 | 1404 | 0.1280 |
| WJM | 109.1054 | -1.11E-16 | 0.5971 | 109.1054 | 0.5890 | 1595 | 0.2425 |
| USM | 7.7857 | -2.22E-16 | 7.7858 | 7.8250 | 0.0625 | 1202 | 0.2029 |
| Grinding | 74.9941 | 0.30001,7 | 74.9939 | 74.9939 | 7.01E-14 | 1506 | 0.1755 |

**Table 9: Results for all problems solved using Modulus Distribution**

| Problem | Function | Constraint | Mean | Best | SD | No. of | Time in |
|---|---|---|---|---|---|---|---|
| AJMB | 8.2540 | 9.66E-05 | 8.2577 | 8.2640 | 0.0044 | 1104 | 0.1147 |
| AJMD | 0.6115 | -7.39E-05 | 0.6068 | 0.6139 | 0.0023 | 1508 | 0.4778 |
| WJM | 136.6380 | -0.017164 | 136.6354 | 136.6380 | 0.0056 | 1126 | 0.1476 |
| USM | 7.9159 | 0.01 | 7.8038 | 7.9896 | 0.2009 | 1102 | 0.1057 |
| Grinding | 75.5091 | 0.3, 6.981 | 75.4790 | 75.5091 | 0.0956 | 1007 | 0.0515 |

**Table 10: Results for all problems solved using Hyperbolic tangent Probability Distribution**

| Problem | Function | Constraint | Mean | Best | SD | No. of | Time in |
|---|---|---|---|---|---|---|---|
| AJMB | 9.4631 | 0.0955 | 9.5172 | 9.8700 | 0.2506 | 1225 | 0.1271 |
| AJMD | 0.6153 | 0.0053 | 0.6170 | 0.6198 | 0.0033 | 1456 | 0.1349 |
| WJM | 136.6380 | -0.0171 | 136.6314 | 136.6380 | 0.0106 | 1602 | 0.2783 |
| USM | 8.05325 | 0.0200 | 8.0465 | 8.0589 | 0.0126 | 1213 | 0.2432 |
| Grinding | 77.5875 | 0.2459,6.9001 | 77.5688 | 77.5906 | 0.0296 | 1456 | 0.1616 |

**12**

Table 11: Abrasive Jet Machining for Brittle materials

| Algorithm | Function Value | Constraint Value | SD | No. of iterations | Time in Seconds |
|---|---|---|---|---|---|
| Triangular | 8.2528 | -4.44E-16 | 2.85E-02 | 1342 | 0.1408 |
| Modulus | 8.2523 | 4.49E-03 | -3.89E-05 | 1104 | 0.1147 |
| Hyperbolic tangent | 9.4631 | 9.55E-02 | 2.50E-01 | 1225 | 0.1271 |
| GA | 8.2423 | -8.48E-04 | NA | 7600 | NA |
| SA | 8.2525 | -2.33E-05 | NA | NA | NA |
| TLBO | 8.7974 | 5.77E-02 | NA | 1000 | NA |
| HH | 8.2481 | -8.61E-05 | 7.94E-08 | 25,500 | 40.2795 |
| COA | 8.2528 | -1.33E-07 | 0 | 240 | 0.8112 |

Note: NA- Not Available

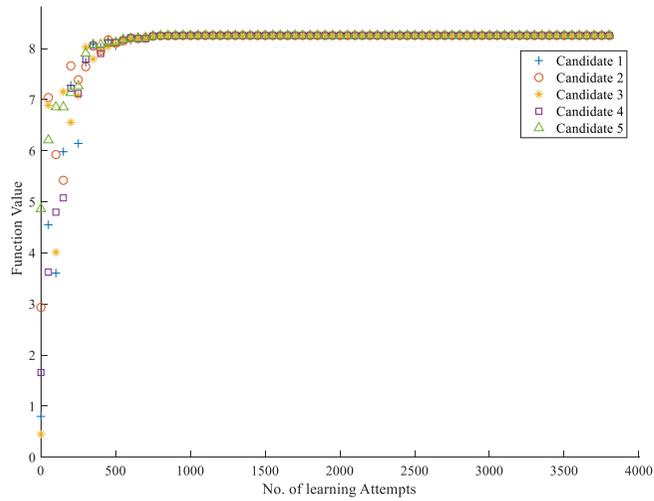

(a) Triangular probability distribution

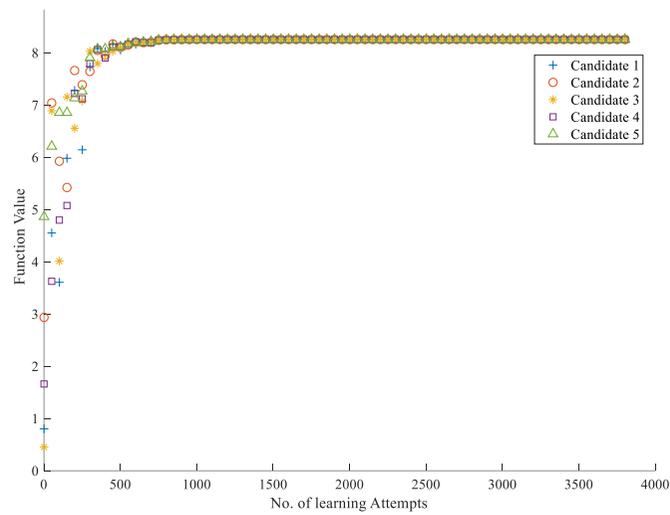

(b) Modulus function

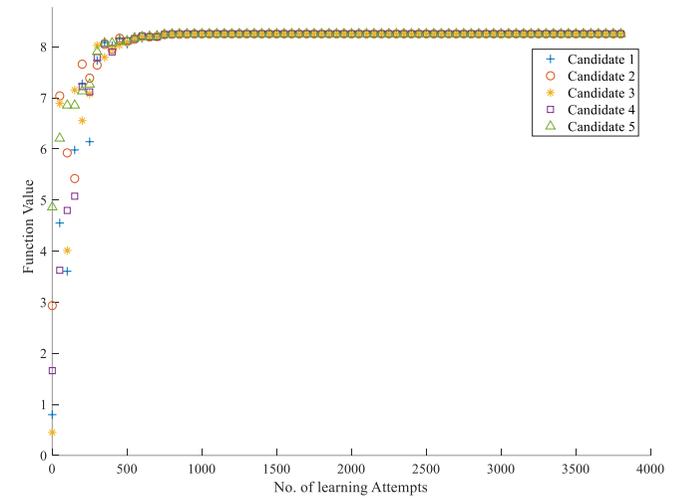

(c) Hyperbolic tangent probability distribution

Figure 7: Convergence Plots for AJM of Brittle Materials

Table 12: Abrasive Jet Machining for Ductile materials

| Algorithm | Function Value | Constraint Value | SD | No. of iterations | Time in Seconds |
|---|---|---|---|---|---|
| Triangular | 0.6056 | -1.11E-16 | 8.19E-03 | 1404 | 0.1280 |
| Modulus | 0.6115 | 2.35E-03 | -7.39E-05 | 1508 | 0.4777 |
| Hyperbolic tangent | 0.6153 | 5.33E-03 | 3.32E-03 | 1456 | 0.1349 |
| GA | 0.6035 | -1.16E-03 | NA | 4600 | NA |
| SA | 0.6053 | -1.54E-04 | NA | NA | NA |
| TLBO | 0.6053 | 6.22E-02 | NA | 1000 | NA |
| HH | 0.6053 | -1.47E-04 | 1.04E-06 | 25000 | 54.8344 |
| COA | 0.6056 | -4.55E-07 | 1.83E-14 | 480 | 1.0296 |

Note: NA- Not Available

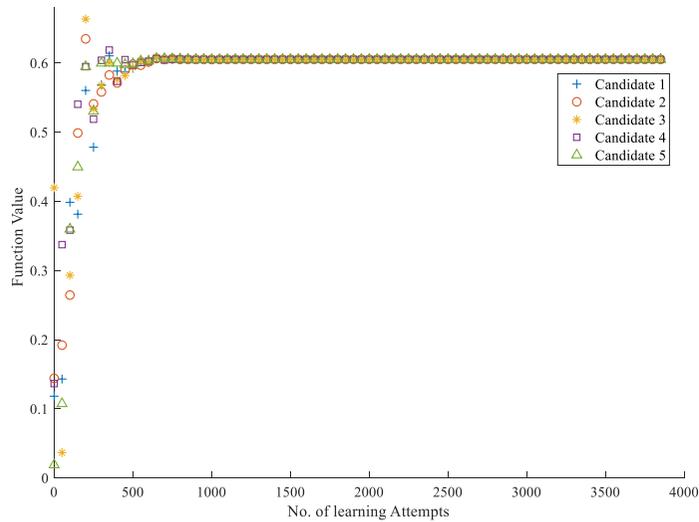 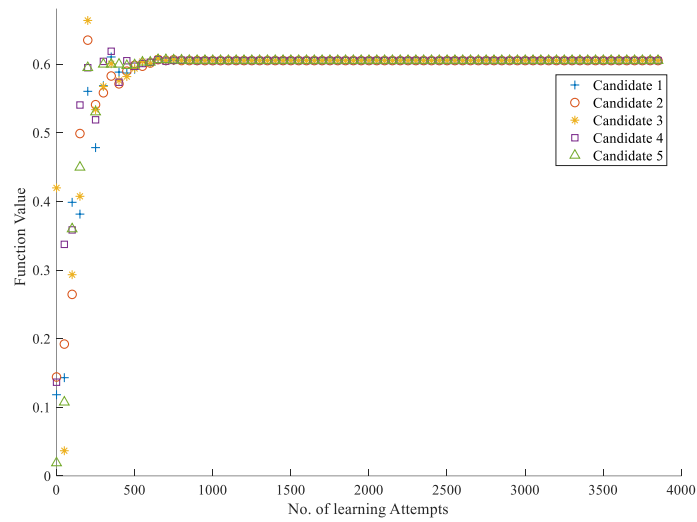 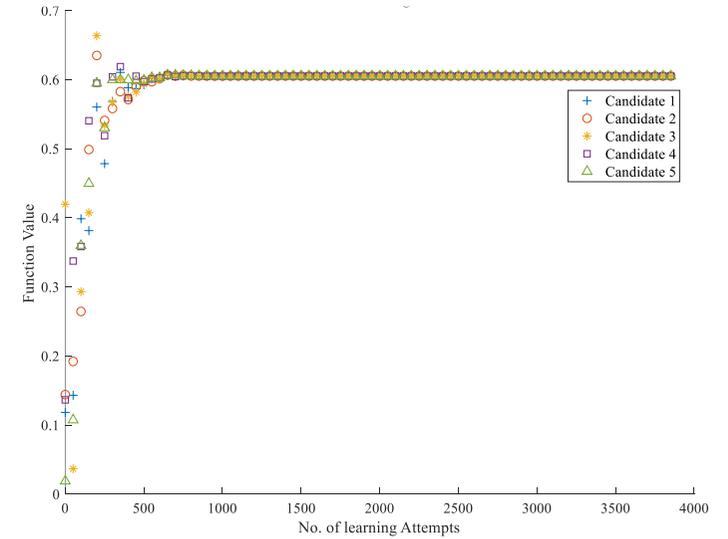

(a) Triangular probability distribution  (b) Modulus function  (c) Hyperbolic tangent probability distribution

Figure 8: Convergence Plots for AJM of Ductile Materials

Table 13: Water Jet machining

| Algorithm | Function Value | Constraint Value | SD | No. of iterations | Time in Seconds |
|---|---|---|---|---|---|
| Triangular | 109.1054 | -1.11E-16 | 0.5890 | 1595 | 0.2425 |
| Modulus | 136.6380 | -0.0171 | 0.0056 | 1126 | 0.1476 |
| Hyperbolic tangent | 136.6380 | -0.0171 | 0.0106 | 1602 | 0.2783 |
| GA | 134.2452 | -0.0282 | NA | 800 | NA |
| HH | 134.3757 | -0.0211 | 3.54E-06 | 1800 | 21.8425 |
| COA | 136.6380 | -0.0171 | 0 | 150 | 0.39 |

Note: NA- Not Available

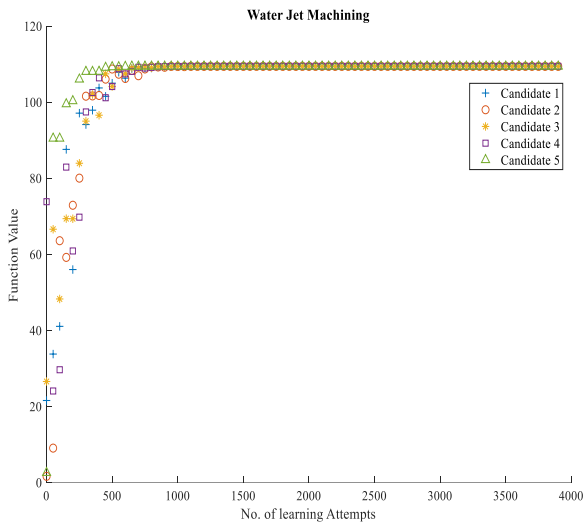
(a) Triangular probability distribution

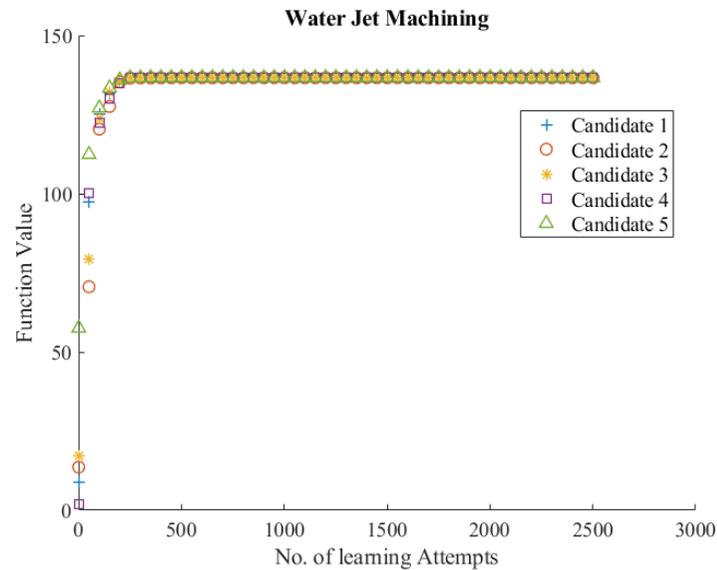
(b) Modulus function

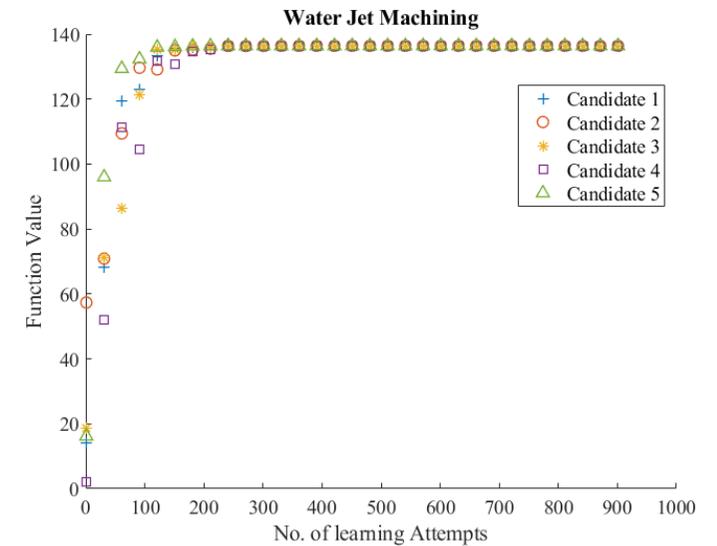
(c) Hyperbolic tangent probability distribution

Figure 9: Convergence Plots for WJM

Table 14: Ultrasonic Machining

| Algorithm | Function Value | Constraint Value | SD | No. of iterations | Time in Seconds |
|---|---|---|---|---|---|
| Triangular | 7.7857 | -2.22E-16 | 0.0625 | 1202 | 0.2029 |
| Modulus | 7.9092 | 0.0070 | 0.2009 | 1102 | 0.1057 |
| Hyperbolic tangent | 8.0532 | 0.0020 | 0.0126 | 1213 | 0.2432 |
| GA | 3.5530 | -0.0214 | NA | 12600 | NA |
| ABC | 3.9410 | -0.0224 | NA | 750 | NA |
| HS_M | 3.8700 | -0.0244 | NA | 750 | NA |
| PSO | 3.9500 | -0.0095 | NA | 250 | NA |
| SA | 3.6600 | -0.0185 | NA | 20000 | NA |
| SFL | 3.894 | -0.0079 | NA | 1000 | NA |
| TLBO | 4.004 | -0.0003 | NA | 1000 | NA |
| HH | 3.7415 | -2.34E-05 | 1.40E-08 | 11700 | 36.9254 |
| COA | 4.0064 | -7.55E-06 | 0 | 225 | 0.7644 |

Note: NA- Not Available

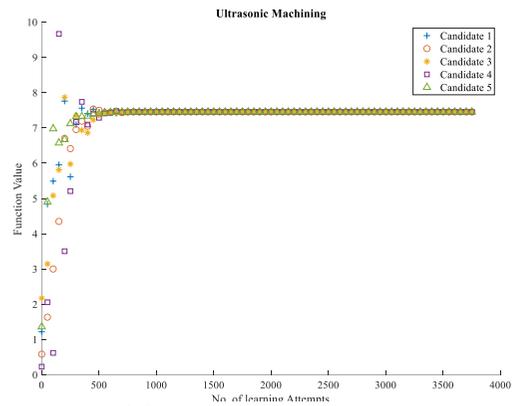 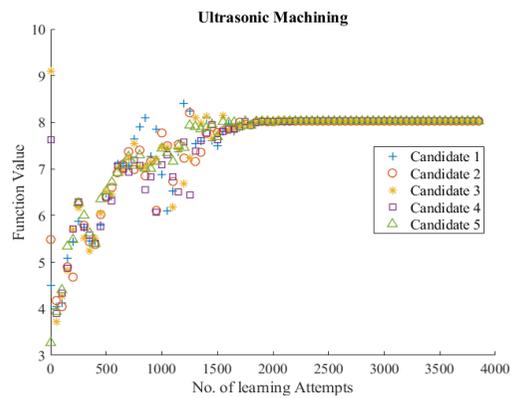 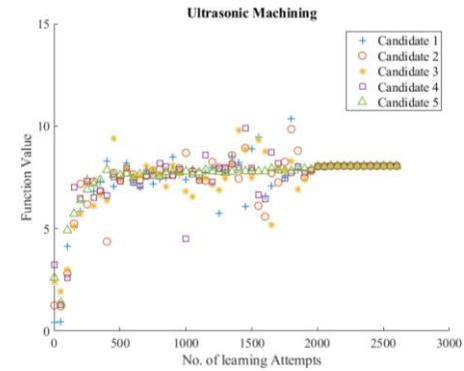

(a) Triangular probability distribution  (b) Modulus function  (c) Hyperbolic Tangent probability distribution

Figure 10: Convergence Plots for USM

Table 15: Grinding

| Algorithm | Function Value | Constraint Value | SD | No. of iterations | Time in Seconds |
|---|---|---|---|---|---|
| Triangular | 74.9939 | 0.3000,7 | 7.01E-14 | 1506 | 0.1755 |
| Modulus | 75.5054 | 0.3000,6.981 | 0.0956 | 1007 | 0.0515 |
| Hyperbolic tangent | 77.5875 | 0.2459,6.9001 | 0.0296 | 1456 | 0.1616 |
| GA | 74.3400 | 0.2990,6.9890 | NA | 9000 | NA |
| GA | 75.1284 | 0.2975,7.0000 | 0.0995 | 10000 | NA |
| PSO | 75.1341 | 0.2959,7.0000 | 7.18E-14 | 10000 | NA |
| NM-PSO | 75.1342 | 0.2986,7.0000 | 2.94E-10 | 6784 | NA |
| ARQiEA | 75.1342 | 0.2959,7.0000 | 1.14E-13 | 3153 | NA |
| DE | 75.1342 | 0.2998,7.0000 | 0.00E+00 | 10000 | NA |
| TLBO | 80.2133 | 0.2993,7.2549 | NA | 8000 | NA |
| HH | 75.1332 | 0.2992,7.0000 | 8.78E-16 | 9000 | 42.9700 |
| COA | 75.1342 | 0.2959,7.0000 | 0.00E+00 | 950 | 1.7472 |

Note: NA- Not Available

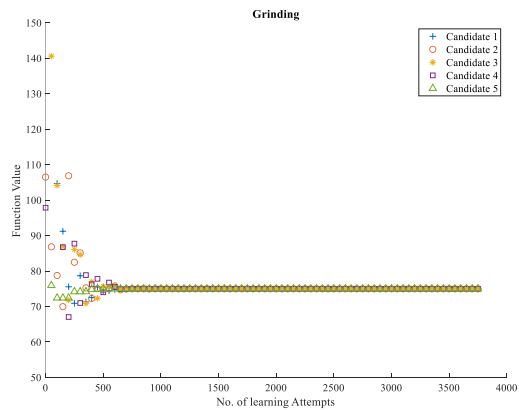
(a) Triangular distribution

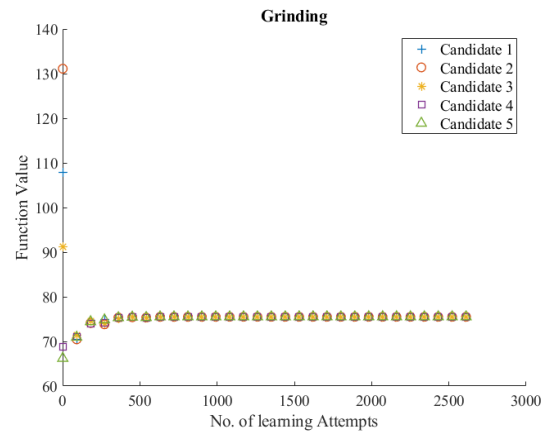
(b) Modulus function

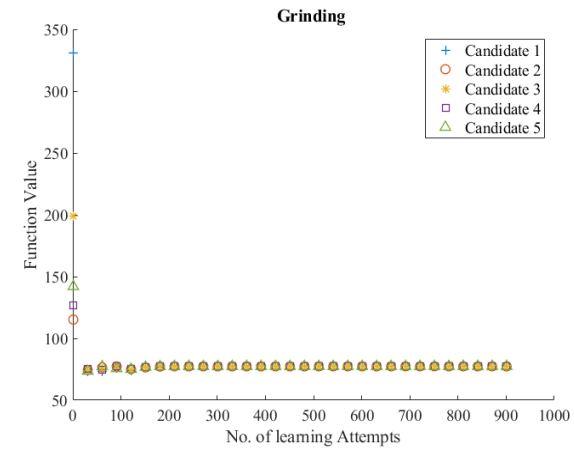
(c) Hyperbolic tangent probability distribution

Figure 11: Convergence Plots for Grinding

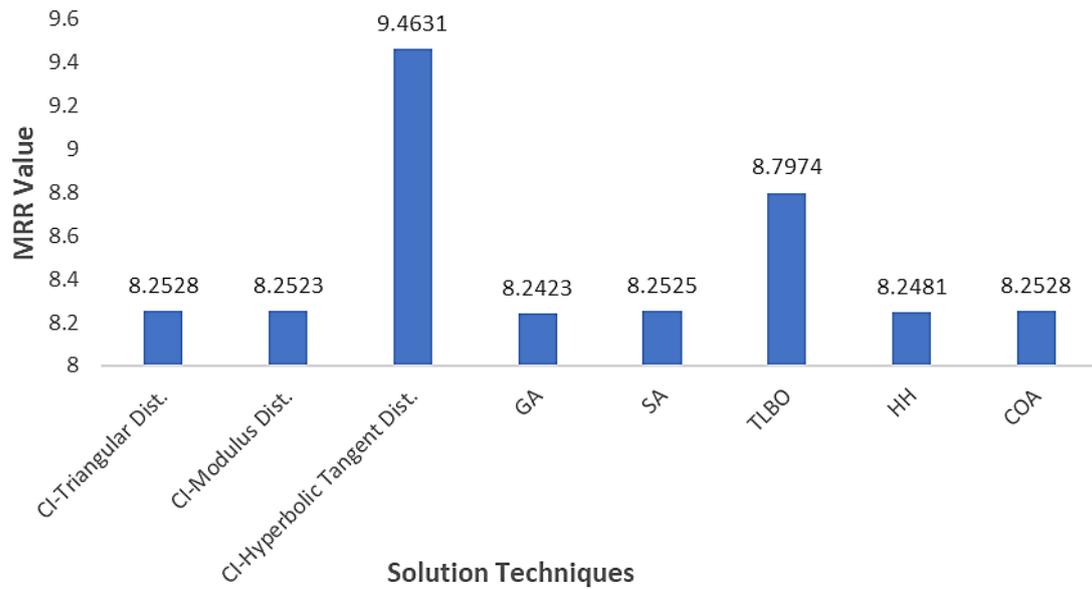
(a) AJM Process for Brittle Materials

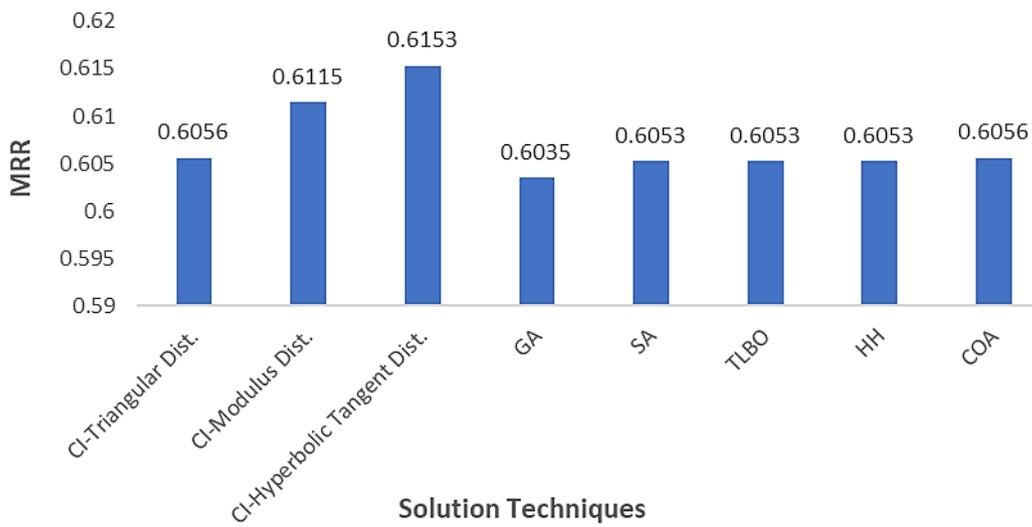
(b) AJM Process for Ductile Materials

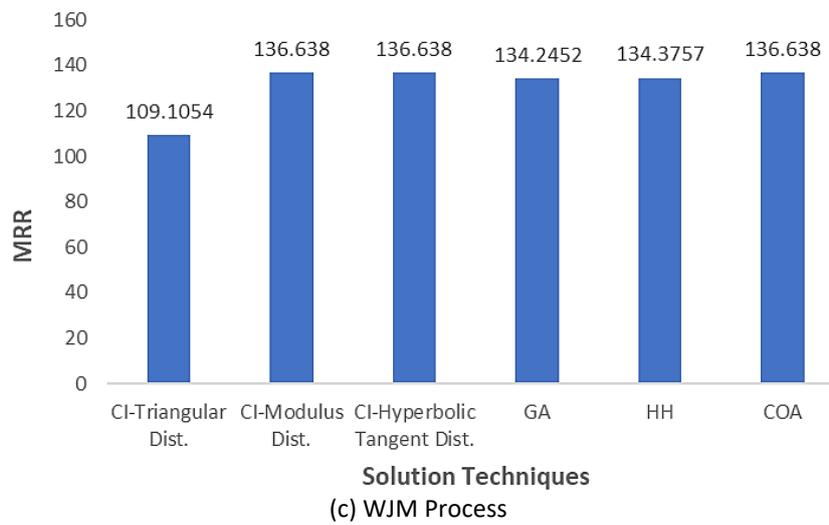
(c) WJM Process

**18**

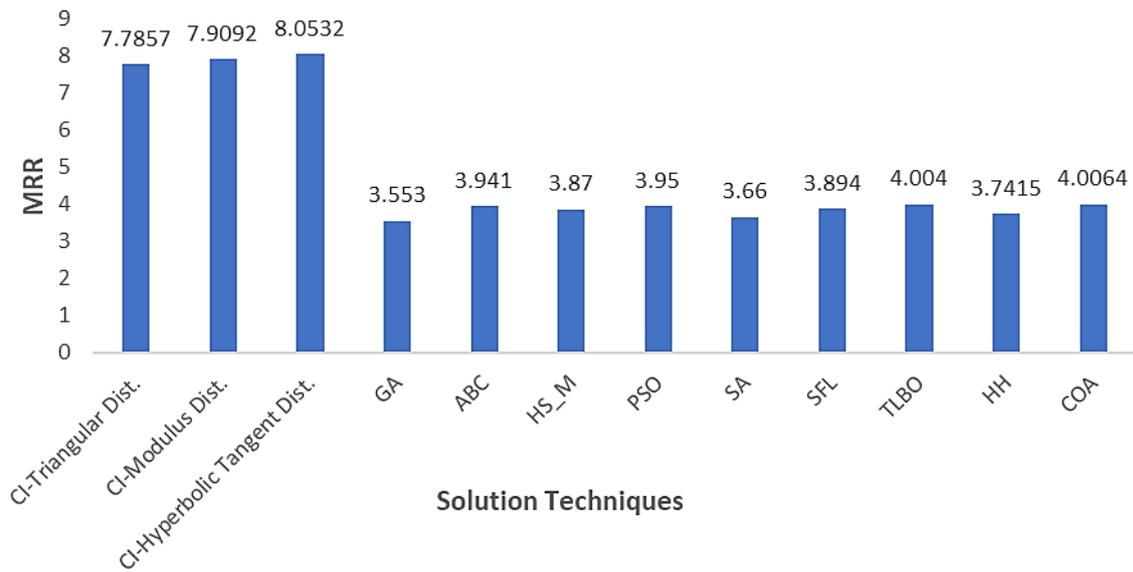

(d) USM Process

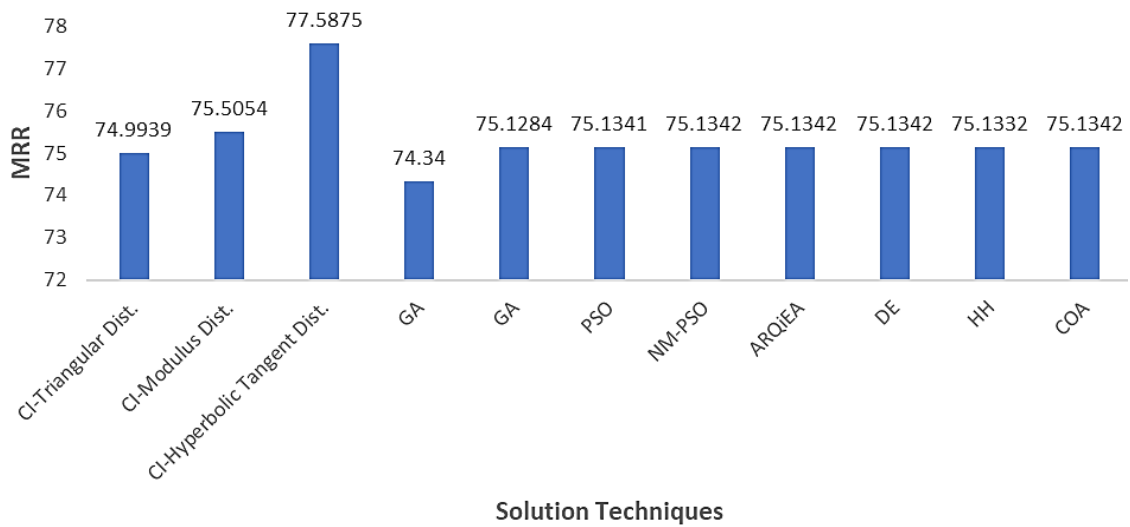

(e) Grinding Process
Figure 12: Comparative Analysis of MRR

Tables 11-15 present the solutions using proposed approaches along with comparison with contemporary algorithms. From Table 11, it is clear that the Modulus function offers a competitive solution, Hyperbolic tangent algorithm gives a 14.6% better solution as compared to the Triangular probability distribution in the maximization of the MRR for AJM with brittle materials. Among the other algorithms, TLBO is superior to GA, SA, HH, COA in terms of solution but is outperformed by the Hyperbolic tangent algorithm that delivers a 7.5% better solution. Hyperbolic tangent method and Modulus methods also supersede HH and COA in terms of computation time as it is 99.6% faster than HH and 85% faster than COA. Table 12 shows the results of the AJM process for ductile materials. It is clear that the Hyperbolic tangent method gives an improvement of 1.6% over the triangular distribution. Similarly, the Modulus method also shows an improvement of 0.97% over the Triangular method and achieves a better result as compared to GA, SA, TLBO, HH and COA. The Hyperbolic tangent method is 99.7% faster than HH and 87% faster than COA. This is also one of the examples of places where the Modulus method, due to its bulky calculations, takes a longer time to arrive at the answer. In Table 13, the results of the WJM problem are presented. It is evident that the Triangular probability distribution fails to achieve the global maximum and converges to a local maximum. The Hyperbolic tangent and Modulus methods are both able to achieve a competitive solution to COA with constraint satisfaction. The Hyperbolic tangent method is 98% faster than HH and 30.7% faster than COA.

Table 14 shows a detailed comparison of the performance of Triangular, Modulus and Hyperbolic tangent probability distribution for solving the advanced machining problem of Ultrasonic Machining. We see that, all the three methods performed exceedingly well, with Hyperbolic tangent giving the best value overall. In comparison to all the other algorithms we see an improvement of up to 127% in the value of Material Removal rate when comparing Hyperbolic tangent distribution method to GA. Modulus probability distribution gives 97.43% better result compared to COA with an 86% less runtime in comparison to the algorithm. The methods also show exceedingly low runtime in comparison to HH algorithm. The Table 15 tabulates the results for Grinding Process. The objective of this problem was to increase the MRR within the constraints of surface roughness and Number of flaws. Through these results we see that Triangular distribution method gives results on par with Genetic Algorithm. However, we observe that Modulus and Hyperbolic tangent distribution methods performed showed competitive results, with Hyperbolic tangent coming on top in comparison to PSO, NM-PSO, DE, HH & COA with up to 4% improved value for MRR. TLBO seems to give the best result out of all the algorithms however, the algorithm breaks the constraint of Number of flaws by 4%.

The shortcoming of the Triangular method is that the degree of extent of constraint violation outside the boundaries of the triangle is not given any weightage. A solution near the boundary point and a solution far away from the previous point would be given the same value of penalty. This can lead to early convergence to a local optimum. This is clearly visible in the problems of WJM and USM. Figures 7-11, a-c presents the convergence plots for AJMB, AJMD, WJM, USM, and Grinding processes with Triangular, Modulus and Hyperbolic tangent probability distributions, respectively. The convergence plots for USM show that the algorithm prefers to converge very fast with the triangular method. This quality combined with the possibility of being misled with a maximum region having the same penalty has a high chance of converging to a wrong solution. This is visible in the grinding problem where the algorithm does not give the result most commonly achieved by other algorithms. The method also fails to achieve the global Maximum and converges at the local Minimum with a significantly less standard deviation. Figure 12 (a-e) presents the comparison of MRR solutions for AJMB, AJMD, WJM, USM, and Grinding processes, respectively in terms of the bar plots. As compared to the GA, CI with Hyperbolic tangent probability distribution achieved 15%, 2%, 2%, 127%, and 4% improvement in MRR for AJMB, AJMD, WJM, USM, and Grinding processes, respectively.

It is evident from Eq. (7-20) that there exists nonlinear relationship and interrelation between the parameters affecting the responses. This makes the problems complex. It is critical in the manufacturing industry to achieve the production target (Takt Time) by maximizing the MRR with required level of surface roughness limiting the power consumption. In the present study, the optimum combinations of parameters for maximizing the MRR are achieved (refer Tables 11-15). The industry can directly adopt these parameters for achieving the desired surface finish, keeping the cutting forces and tool wear minimum. The problems solved in the paper has the objective of maximizing the MRR, the significantly higher result will greatly help in cutting down the time of machining in turn improving the productivity. The deviation of the constraint values in the order of E-01 and less would not greatly impact the finish of the workpiece. Moreover, it is the requirement of the customer that would decide the finish of the workpiece. Therefore, if the demand sates that such a deviation is fine in the surface finish, the values obtained by the proposed method would prove to be cheaper and better.

## 5. Conclusions & Future Directions
In this paper, a constrained CI algorithm with constraint handling approaches based on three probability distributions viz. triangular, modulus and Hyperbolic tangent is presented. In the existing triangular based probability distribution approach, the probability of the constraint values outside the set constraints limits is more than that at the limits. This makes the points outside the constraint limits favorable which may cause the algorithm to go awry. Moreover, the points closer to the limits but outside the distribution triangle and the points far away from the triangle, both are assigned the same amount of penalty. This reduces the exploration capabilities of the algorithm. The resulting values in this case may satisfy constraints but could be local optima. Hence, two constraint handling approaches based on modulus function and Hyperbolic tangent distribution are developed. The proposed approaches are validated by solving selected three constrained test problems viz. G1, G4 and G6.
Furthermore, real world applications of proposed approaches are also successfully demonstrated in the advanced manufacturing domain by solving nonlinear, non-separable and multimodal problems for optimizing four advanced manufacturing processes such as AJM, WJM, USM and Grinding. The problems solved are maximization of $MRR$ for USM and AJM with brittle & ductile materials considering the $R_a$ constraint. Furthermore, maximization of $MRR$ for WJM & Grinding with constraints such as power usage for WJM and $R_a$

& no of flaws for Grinding processes are also solved. The performance of proposed approaches is validated by comparing the results with several contemporary algorithms such as GA, SA, PSO, TLBO, HH, COA, etc. The proposed approaches achieved 1.6%-127% maximization of MRR satisfying all hard constraints for all the problems. As compared to the GA, CI with Hyperbolic tangent probability distribution achieved 15%, 2%, 2%, 127%, and 4% improvement in MRR for AJMB, AJMD, WJM, USM, and Grinding processes, respectively.

The analysis regarding the convergence of all the algorithms is discussed in detail. For most of the problems, the run time for the proposed CI is lower as compared to other algorithms. In addition, the standard deviation is comparatively much lower which exhibited its robustness. This demonstrates the exploration and exploitation capabilities of the proposed approaches. The major challenge encountered during experimental validation is parameter settings. There are certain parameters such as no. of candidates, sample space reduction factor that necessarily drive the performance of the algorithm. These parameters are required to be set using preliminary trials.

The contributions in this paper have opened several avenues for further applicability of the proposed constraint handling approaches for solving complex constrained problems. In the near future, authors intend to apply the proposed approaches for solving Multi-Objective Optimization problems adopting goal programming formulations associated with a variety of domains including supply-chain management, transportation, dynamic control problems, etc.

**Competing Interests**
Authors have no competing/conflicting interests of any kind.

**Funding Information**
No funding was received for conducting this study.

**References**

1. Ahmadi-Javid, A., & Hooshangi-Tabrizi, P. (2017). Integrating employee timetabling with scheduling of machines and transporters in a job-shop environment: A mathematical formulation and an Anarchic Society Optimization algorithm. Computers & Operations Research, 84, 73-91.
2. Arora, J. S. (2004). Introduction to optimum design. Elsevier.
3. Askarzadeh, A. (2016). A novel metaheuristic method for solving constrained engineering optimization problems: crow search algorithm. Computers & Structures, 169, 1-12
4. Biswas, J. H., Jagadish, & Ray, A. (2019). Experimental investigation and optimization of ultrasonic machining parameters on zirconia composite. International Journal of Machining and Machinability of Materials, 21(1-2), 115-137.
5. Biswas, P. P., Suganthan, P. N., Mallipeddi, R., & Amaratunga, G. A. (2018). Optimal power flow solutions using differential evolution algorithm integrated with effective constraint handling techniques. Engineering Applications of Artificial Intelligence, 68, 81-100.
6. Braune, R., Wagner, S., & Affenzeller, M. (2004). Applying genetic algorithms to the optimization of production planning in a real-world manufacturing environment.
7. Chen, H., Wang, T., Chen, T., & Deng, W. (2023). Hyperspectral Image Classification Based on Fusing S3-PCA, 2D-SSA and Random Patch Network. Remote Sensing, 15(13), 3402.
8. Cheng, P., Wang, H., Stojanovic, V., Liu, F., He, S., & Shi, K. (2022). Dissipative-based finite-time asynchronous output feedback control for wind turbine system via a hidden Markov model. International Journal of Systems Science, 53(15), 3177-3189
9. Das, S., Doloi, B., & Bhattacharyya, B. (2013). Optimization of ultrasonic machining of zirconia bio-ceramics using genetic algorithm. International Journal of Manufacturing Technology and Management 4 and 25, 27(4-6), 186-197.
10. Deb, K. (2012). Optimization for engineering design: Algorithms and examples. PHI Learning Pvt. Ltd.
11. Djordjevic, V., Tao, H., Song, X., He, S., Gao, W., & Stojanović, V. (2023). Data-driven control of hydraulic servo actuator: An event-triggered adaptive dynamic programming approach. Mathematical biosciences and engineering, 20(5): 8561–8582
12. Du DZ., Pardalos P.M., Wu W. (2008) History of Optimization. In: Floudas C., Pardalos P. (eds) Encyclopedia of Optimization. Springer, Boston, MA
13. Emami, H., & Derakhshan, F. (2015). Election algorithm: A new socio-politically inspired strategy. AI Communications, 28(3), 591-603.
14. Gandomi, A. H., Yang, X. S., & Alavi, A. H. (2013). Cuckoo search algorithm: a metaheuristic approach to solve structural optimization problems. Engineering with computers, 29(1), 17-35.
15. Goldberg, D. E. (1989). Genetic algorithms for search, optimization, and machine learning. Reading, MA: Addison-Wesley
16. Goswami, D., & Chakraborty, S. (2015). Parametric optimization of ultrasonic machining process using gravitational search and fireworks algorithms. Ain Shams Engineering Journal, 6(1), 315-331.
17. Gulia, V., & Nargundkar, A. (2019). Optimization of process parameters of abrasive water jet machining using variations of cohort intelligence (CI). In Applications of Artificial Intelligence Techniques in Engineering: SIGMA 2018, Volume 2 (pp. 467-474). Springer Singapore.
18. Hatamlou, A. (2013). Black hole: A new heuristic optimization approach for data clustering. Information sciences, 222, 175-184.
19. He, P., Wen, J., Stojanovic, V., Liu, F., & Luan, X. (2022). Finite-time control of discrete-time semi-Markov jump linear systems: A self-triggered MPC approach. Journal of the Franklin Institute, 359(13), 6939-6957
20. He, X. S., Fan, Q. W., Karamanoglu, M., & Yang, X. S. (2019, June). Comparison of constraint-handling techniques for metaheuristic optimization. In International Conference on Computational Science (pp. 357-366). Springer, Cham.
21. Homaifar A, S.H.Y. Lai, X. Qi, Constrained optimization via genetic algorithms, Simulation 62 (4) (1994) 242–254.
22. Huan, T. T., Kulkarni, A. J., Kanesan, J., Huang, C. J., & Abraham, A. (2017). Ideology algorithm: a socio-inspired optimization methodology. Neural Computing and Applications, 28(1), 845-876.
23. Huyer. W., Neumair A, A new exact penalty function, SIAM Journal of Optimization, 13 (2003) 1141–1159.

24. Jain, N. K., Jain, V. K., & Deb, K. (2007). Optimization of process parameters of mechanical type advanced machining processes using genetic algorithms. International Journal of Machine Tools and Manufacture, 47(6), 900-919
25. Jain, V.K., 2008. Advanced (non-traditional) machining processes. In Machining (pp. 299-327). Springer, London.
26. Joines J, Houck C, "On the use of non-stationary penalty functions to solve nonlinear constrained optimization problems with Gas", in: D. Fogel (Ed.), Proceedings of the First IEEE Conference on Evolutionary Computation, IEEE Press, Orlando, FL, 1994, pp. 579–584.
27. Jones, D., & Tamiz, M. (2010). Practical goal programming (Vol. 141). New York: Springer.
28. Kale, I. R., & Kulkarni, A. J. (2021). Cohort intelligence with self-adaptive penalty function approach Hybridized with colliding bodies optimization algorithm for discrete and mixed variable constrained problems. Complex & Intelligent Systems, 1-32.
29. Kashan, A. H. (2009, December). League championship algorithm: a new algorithm for numerical function optimization. In 2009 International Conference of Soft Computing and Pattern Recognition pp. 43-48.
30. Kennedy, J., & Eberhart, R. (1995, November). Particle swarm optimization (PSO). In Proc. IEEE International Conference on Neural Networks, Perth, Australia pp. 1942-1948.
31. Kulkarni, A. J., & Shabir, H. (2016). Solving 0–1 knapsack problem using cohort intelligence algorithm. International Journal of Machine Learning and Cybernetics, 7(3), 427-441.
32. Kulkarni, A. J., Durugkar, I. P., & Kumar, M. (2013). Cohort Intelligence: A Self Supervised Learning Behavior. 2013 IEEE International Conference on Systems, Man, and Cybernetics.
33. Kulkarni, A.J., Tai, K., Abraham, A.: Probability collectives: a distributed multi-agent system approach for optimization. In: Intelligent Systems Reference Library, vol. 86. Springer, Berlin (2015)
34. Kulkarni, O., Kulkarni, N., Kulkarni, A. J., & Kakandikar, G. (2018). Constrained cohort intelligence using static and dynamic penalty function approach for mechanical components design. International Journal of Parallel, Emergent and Distributed Systems, 33(6), 570-588.
35. Kumar, K. V., Suryakumari, T. S. A., & Mohanavel, V. (2020). A Review on methods used to optimize Abrasive Jet Machining parameters. Materials Today: Proceedings.
36. Kuo, H. C., & Lin, C. H. (2013). Cultural evolution algorithm for global optimizations and its applications. Journal of applied research and technology, 11(4), 510-522.
37. Lagaros, N.D., Kournoutos, M., Kallioras, N.A., Nordas, A.N. Constraint handling techniques for metaheuristics: a state-of-art review and new variants, Optimization and Engineering, 2023.
38. Li, X., Zhao, H., & Deng, W. (2023). BFOD: Blockchain-based privacy protection and security sharing scheme of flight operation data. IEEE Internet of Things Journal.
39. Liu, Z. Z., Chu, D. H., Song, C., Xue, X., & Lu, B. Y. (2016). Social learning optimization (SLO) algorithm paradigm and its application in QoS-aware cloud service composition. Information Sciences, 326, 315-333.
40. Mallipeddi, R., & Suganthan, P. N. (2010). Ensemble of constraint handling techniques. IEEE Transactions on Evolutionary Computation, 14(4), 561-579.
41. Mellal, M. A., & Williams, E. J. (2016). Parameter optimization of advanced machining processes using cuckoo optimization algorithm and hoopoe heuristic. Journal of Intelligent Manufacturing, 27(5), 927-942.
42. Michalewicz Z, Attia N.F, "Evolutionary optimization of constrained problems", in: Proceedings of the 3rd Annual Conference on Evolutionary Programming, World Scientific, Singapore, 1994, pp. 98–108.
43. Mishra P K, Noncoventional machining, in: The Institution of Engineers (India), 2012, pp. 45–52, ISBN: 978-81-7319-138-1.
44. Moosavian, N. (2015). Soccer league competition algorithm for solving knapsack problems. Swarm and Evolutionary Computation, 20, 14-22.
45. Pansari, S., Mathew, A., & Nargundkar, A. (2019). An investigation of burr formation and cutting parameter optimization in micro-drilling of brass C-360 using image processing. In Proceedings of the 2nd International Conference on Data Engineering and Communication Technology: ICDECT 2017 (pp. 289-302). Springer Singapore.
46. Patankar, N. S., & Kulkarni, A. J. (2018). Variations of cohort intelligence. Soft Computing, 22(6), 1731-1747.
47. Ponsich, A., Azzaro-Pantel, C., Domenech, S., & Pibouleau, L. (2008). Constraint handling.

48. Rahimi, I., Gandomi, A. H., Chen, F., & Mezura-Montes, E. (2022). A Review on Constraint Handling Techniques for Population-based Algorithms: from single-objective to multi-objective optimization. Archives of Computational Methods in Engineering, 1-29.
49. Rao, R. V. (2011). Modeling and optimization of modern machining processes. In Advanced modeling and optimization of manufacturing processes. (ch. 3, pp. 177–284). London: Springer
50. Rao, R. V. (2016). Teaching-learning-based optimization algorithm. In Teaching learning based optimization algorithm (pp. 9-39). Springer, Cham.
51. Rao, R. V., & Kalyankar, V. D. (2014). Optimization of modern machining processes using advanced optimization techniques: a review. The International Journal of Advanced Manufacturing Technology, 73(5-8), 1159-1188.
52. Shastri, A. S., & Kulkarni, A. J. (2018). Multi-cohort intelligence algorithm: an intra-and inter-group learning behaviour based socio-inspired optimisation methodology. International Journal of Parallel, Emergent and Distributed Systems, 33(6), 675-715.
53. Shastri, A. S., Jadhav, P. S., Kulkarni, A. J., & Abraham, A. (2016). Solution to constrained test problems using cohort intelligence algorithm. In Innovations in Bio-Inspired Computing and Applications (pp. 427-435). Springer, Cham.
54. Shastri, A. S., Nargundkar, A., Kulkarni, A. J., & Sharma, K. K. (2020). Multi-cohort intelligence algorithm for solving advanced manufacturing process problems. Neural Computing and Applications, 32(18), 15055-15075.
55. Shastri, A., Nargundkar, A., & Kulkarni, A. J. (2021). Socio-Inspired Optimization Methods for Advanced Manufacturing Processes (pp. 19-29). Singapore:: Springer.
56. Shastri, A., Nargundkar, A., Kulkarni, A. J., & Benedicenti, L. (2021). Optimization of process parameters for turning of titanium alloy (Grade II) in MQL environment using multi-CI algorithm. SN Applied Sciences, 3(2), 1-12.
57. Shastri, A.S., Thorat, E.V., Kulkarni, A.J. and Jadhav, P.S., 2019, Optimization of Constrained Engineering Design Problems Using Cohort Intelligence Method. In Proceedings of the 2nd International Conference on Data Engineering and Communication Technology (pp. 1-11). Springer, Singapore
58. Simon, D. (2008). Biogeography-based optimization. IEEE transactions on evolutionary computation, 12(6), 702-713.
59. Song, Y., Zhao, G., Zhang, B., Chen, H., Deng, W., & Deng, W. (2023). An enhanced distributed differential evolution algorithm for portfolio optimization problems. Engineering Applications of Artificial Intelligence, 121, 106004.
60. Tomy, A., & Hiremath, S. S. (2020). Machining and Characterization of Multidirectional Hybrid Silica Glass Fiber Reinforced Composite Laminates Using Abrasive Jet Machining. Silicon, 1-14.
61. Van Laarhoven, P. J., & Aarts, E. H. (1987). Simulated annealing. In Simulated annealing: Theory and applications (pp. 7-15). Springer, Dordrecht.
62. Zhao, H., Liu, J., Chen, H., Chen, J., Li, Y., Xu, J., & Deng, W. (2022). Intelligent diagnosis using continuous wavelet transform and gauss convolutional deep belief network. IEEE Transactions on Reliability.